\documentclass[letterpaper,final]{article}

\PassOptionsToPackage{authoryear,square,semicolon}{natbib}
\usepackage{optrl_2019}

\usepackage[utf8]{inputenc} 
\usepackage[T1]{fontenc}    
\usepackage{newtxmath}
\usepackage[scaled=0.95]{cabin}

\usepackage[hyphens]{url} 
\usepackage[pdfusetitle]{hyperref}
\usepackage[hyphenbreaks]{breakurl}
\usepackage[capitalize,nameinlink,noabbrev]{cleveref}
\usepackage{doi}

\hypersetup{
  bookmarksnumbered,unicode,psdextra,
  colorlinks=true,allcolors=black,urlcolor=blue,breaklinks,
  pdfauthor={Abhishek Naik, Roshan Shariff, Niko Yasui, Hengshuai Yao, Richard S. Sutton}
}

\usepackage{amsmath}
\usepackage{mathtools}
\usepackage{textcomp}       
\usepackage{amsfonts}       
\usepackage{nicefrac}       
\usepackage{microtype}      
\usepackage{booktabs}       
\usepackage{enumitem}       
\usepackage{subcaption}
\usepackage{multicol}

\usepackage[obeyFinal,textsize=small]{todonotes}

\usepackage{tikz}
\usetikzlibrary{chains,arrows}

\title{Discounted Reinforcement Learning\\Is Not an Optimization Problem}

\author{%
  Abhishek Naik\textsuperscript{\normalfont 1,2}\thanks{This work was partially done during an internship at Huawei Technologies.}\quad
  Roshan Shariff\textsuperscript{\normalfont 1}\quad
  Niko Yasui\textsuperscript{\normalfont 1}\quad
  Hengshuai Yao\textsuperscript{\normalfont 2}\quad
  Richard S. Sutton\textsuperscript{\normalfont 1}\\
  \texttt{\{anaik1,rshariff,yasui,rsutton\}@ualberta.ca\quad
  hengshuai.yao@huawei.com}%
}

\DeclareMathOperator*{\argmax}{arg\,max}

\begin{document}

\maketitle

\vspace{-1.82\baselineskip}
\begin{center}
  \begin{tabular}{c@{\qquad}c}
    \llap{\textsuperscript{1}}Department of Computing Science
    & \llap{\textsuperscript{2}}Huawei Technologies\\
    University of Alberta
    & Edmonton, Canada
  \end{tabular}
\end{center}
\vspace{1\baselineskip}
\setcounter{footnote}{0}

\begin{abstract}
    Discounted reinforcement learning is fundamentally incompatible with function approximation for control in continuing tasks. It is not an optimization problem in its usual formulation, so when using function approximation there is no optimal policy. We substantiate these claims, then go on to address some misconceptions about discounting and its connection to the average reward formulation. We encourage researchers to adopt rigorous optimization approaches, such as maximizing average reward, for reinforcement learning in continuing tasks.
\end{abstract}


\section{Introduction}

Reinforcement learning (RL) is a paradigm in which an agent learns to interact with an environment in order to maximize reward \citep{RLtextbook}. Many interesting RL problems concern continuing tasks in which the agent lives and learns over a single lifetime, rather than experiencing a sequence of distinct episodes. Some examples of continuing tasks include routing internet packets, managing the inventory in a warehouse, and controlling the temperature in a data center.

In continuing tasks, it is common practice to value immediate rewards more highly than rewards further in the future --- this is called temporal discounting.  One reason is that the sum of future rewards (which is finite for episodic tasks) can grow to infinity as the agent and environment continually interact.  With discounting the sum of the infinitely many future rewards remains bounded even when the length of the interaction does not.\footnote{The notion of discounting discussed here is one that leads to a finite sum of future rewards. This occurs with geometric discounting, for instance, but not hyperbolic discounting \citep{Fedus19_HyperbolicDisc}, which is applicable only in finite-horizon episodic problems.}

In this paper, we take the position that this common practice has serious conceptual flaws and should be carefully reconsidered. The straightforward formulation of discounted reinforcement learning is fundamentally incompatible with large-scale reinforcement learning in continuing tasks. Moreover, these issues are inherent to the very idea of discounting and cannot be avoided by minor modifications of the problem formulation.

For continuing tasks, we will first argue that \emph{discounted reinforcement learning is not an optimization problem}, and that this makes it incompatible with function approximation. A standard optimization problem is defined by a set of feasible solutions and an objective function that describes the quality of each feasible solution with a real number. The feasible solutions in RL are policies and, at least for episodic tasks, the natural objective function is the sum of rewards over an episode.   With continuing tasks, however, optimality under discounting is not defined as maximizing an objective function.  Instead, a policy is usually considered optimal if, in every state, it achieves a higher discounted sum of future rewards than any other policy --- for a fixed discount rate $\gamma$, an optimal policy $\pi^*$ satisfies
\begin{align}
\label{eq:opt-policy}
    v_{\pi^*}^\gamma(s) &\ge v_{\pi}^\gamma(s), &\text{for all states $s$ and policies $\pi$}.
\end{align}
These inequalities produce a \emph{partial order} on the set of policies: some pairs of policies may be incomparable with each other, because each achieves a higher value in some states but a lower value in others. This is not a problem with tabular representations, which can represent any possible policy --- in their foundational work, \citet{Bellman59_DP} proved the existence of a policy that maximizes value at every state simultaneously. With function approximation, on the other hand, a partial ordering is not enough to identify an optimal policy.

\enlargethispage{\baselineskip}

\subsection{There is no optimal representable policy with discounting and function approximation}

In many RL problems the state or action spaces are so large that policies cannot be represented as a table of action probabilities for each state. In such domains we often resort to a compact policy representation that cannot represent every possible policy. In most cases the optimal policy defined by~\eqref{eq:opt-policy} will no longer be representable, so the agent cannot hope to learn it. Instead, the agent should find the best representable policy.
 
However, without an objective function there is usually \emph{no} representable policy that is unambiguously better than all the other representable policies.  In general, for every representable policy there will be another policy that has a higher value in some states but a lower value in others, so the partial order will not be able to identify any policy as optimal --- the notion of ``best representable policy'' is not well-defined.  See \citet{Singh94_POMDP_AvgRew} for illustrative examples.

If we had an explicit objective function we would be able to compare any two policies, creating a total order over the policy space.  Regardless of the choice of policy representation, the total ordering provided by an explicit objective function would guarantee the existence of an optimal representable policy.  This is why there is no issue with defining optimal policies for episodic tasks, where the sum of rewards is a natural and explicit objective function.

In the rest of this paper we will address the question of choosing an appropriate objective function for continuing tasks.  We will find that discounting is hard to incorporate into a meaningful objective function.  Many common choices do not capture the continuing nature of the task, whereas others are formally equivalent to undiscounted objectives. Indeed, we will be forced to conclude that discounting is not just a technical problem with the above definition of optimality. Rather, there is a fundamental mismatch between continuing tasks and discounting that cannot be ignored when function approximation is required.

We now argue that function approximation and continuing tasks are indispensable for reinforcement learning, which makes this problem setting important.

\subsection{Continuing control with function approximation is the problem that matters for AI}

Reinforcement learning with tabular representations has served us well to build intuition, construct algorithms, and analyze their properties, but the world we live in is too large to be represented by tables of values or action probabilities. Furthermore, it is crucial to generalize across similar states and actions. Function approximation serves both these purposes and is necessary for agents to compactly represent complex worlds and make sense of their complexity.

Continuing tasks are those in which the agent-environment interaction does not naturally break down into episodes. Lifelong learning follows this paradigm, because an agent living and learning over a lifetime does not usually have the ability to reset to a pre-defined initial state. Additionally, many interesting RL tasks being studied today involve extremely long episodes and the challenges that come with them, such as sparse rewards and credit assignment over long time horizons. Such tasks are episodic in name only, and should be thought of as continuing tasks in spirit. In particular, solutions that fundamentally rely on episodes are likely to fare worse than those that fully embrace the continuing task setting.

Function approximation is necessary in large-scale reinforcement learning, whereas discounting is not. We believe that whole-heartedly adopting optimization approaches for control in continuing tasks (for example, average reward) is the most appropriate direction for our discipline.

\subsection{Paper organization}

After covering some background in \cref{sec:background}, we describe an alternative continuing task formulation in \cref{sec:avg-ref-objective} which has a well-defined objective --- maximizing the average reward per step --- making it suitable for use with function approximation. We discuss its equivalence with the time-average of the discounted value function, and the resulting misconception that common discounted algorithms maximize the average reward irrespective of the choice of the discount factor. We summarize the arguments in \cref{sec:conclusions} and give pointers to the existing literature involving the average reward formulation.


\section{Background}%
\label{sec:background}

A Markov decision process (MDP) consists of a finite set of states $\mathcal{S}$ and a finite set of actions $\mathcal{A}$ that can be performed in those states. When the agent takes an action in a state, the agent transitions to another state and receives a scalar reinforcement signal called the reward. Formally, an MDP can be represented by $M = (\mathcal{S}, \mathcal{A}, P, r, \mu_0)$, where upon performing action $a$ in state $s$ the agent receives reward $r(s,a)$ in expectation and transitions to state $s'$ with probability $P_a(s,s')$. Note that the transition probabilities and rewards are independent of the agent's history before reaching state $s$ (the Markov assumption). The agent's initial state is distributed according to $\mu_0$.
 
 A policy $\pi: \mathcal{S} \times \mathcal{A} \to [0,1]$ gives, for every state in an MDP, a probability distribution over actions to be taken in that state.  A policy $\pi$ acting in an MDP $(\mathcal{S}, \mathcal{A}, P, r, \mu_0)$ produces a Markov reward process $(\mathcal{S}, P_\pi, r_\pi, \mu_0)$, where $P_\pi:\mathcal{S}\times\mathcal{S}\to[0,1]$ is the transition probability and $r_\pi:\mathcal{S}\to\mathbb{R}$ is the reward function:
 \begin{align*}
     P_\pi(s, s') &= \sum_{a\in\mathcal{A}} \pi(s, a) \, P_a(s, s'), \\
     r_\pi(s) &= \sum_{a\in\mathcal{A}} \pi(s, a) \, r(s, a).
 \end{align*}

\subsection*{Average reward}%
\label{sec:avg-reward}

The average reward of a policy is an intuitively straightforward quantity --- it is the reward the agent receives on average every time step as it acts in an environment. We define the average reward more formally as follows. Under certain mild conditions, a Markov reward process has a so-called \emph{stationary distribution} over states, $d_\pi(s)$, that satisfies
\begin{align*}
    d_\pi(s') &= \sum_{s\in\mathcal{S}} d_\pi(s) \, P_\pi(s, s') &&\text{for every }s'\in\mathcal{S}.
\end{align*}
In other words, if the agent's state is distributed according to the stationary distribution and it acts according to its policy, then its next state will also follow the stationary distribution. Furthermore, under additional mild conditions, the state distribution converges to $d_\pi(s)$ over the long run regardless of the starting state $s_0$:
\begin{align*}
    d_\pi(s) &= \lim_{T\to\infty} \Pr(S_T = s), &\text{where } S_0 = s_0 \text{ and }
    S_{t+1} \sim P_\pi(S_t, \,\cdot\,)\text{ for all time steps }t.
\end{align*}
In continuing tasks, $d_\pi(s)$ measures the fraction of time the agent spends in state $s$. Frequently visited states have higher values of $d_\pi$, and if $d_\pi(s)=0$ then $s$ is called a \emph{transient state}: it is only visited a finite number of times and never again.

The average reward of a policy is simply defined as the average one-step reward, weighted by the proportion of time spent in each state while following that policy:
\begin{align}\label{eq:obj-avg-rew}
    \bar{r}(\pi) &= \sum_{s\in\mathcal{S}} d_\pi(s) \, r_\pi(s).
\end{align}

 For the purposes of this paper, we only consider stationary Markov policies, which do not change with time and depend only on the current state, not previous history. For more details on MDPs and average reward, refer to \citet{Puterman94_MDPbook}.


\section{An Optimization Objective for Continuing Tasks}%
\label{sec:avg-ref-objective}

We would like to find an objective for continuing tasks that is analogous to the sum of rewards in episodic tasks. Such an objective should be a function that quantifies the performance of each policy by a single number, so that any policy can be compared with any other and our goal of finding the best policy amongst any class of policies becomes well-defined.

The founding principle of reinforcement learning is that an agent acts to maximize reward not just in the present, but also over its future lifetime, which is potentially infinite in continuing tasks.  The optimality definition of~\eqref{eq:opt-policy} captures this idea --- $v_\pi^\gamma(s)$ represents the future reward (though not infinitely far in the future) which the agent seeks to maximize at every state $s$.

A natural objective function, therefore, is the weighted average of $v_\pi^\gamma(s)$ with each state weighted by $\mu(s)$ --- maximizing $\sum_s\mu(s)\,v_\pi^\gamma(s)$. The choice of $\mu$ determines which states the agent prefers when it cannot act optimally in every state simultaneously. We could contemplate evaluating policies using their discounted value from the start state (setting $\mu=\mu_0$, the initial state distribution), but this gives undue importance to the early part of an agent's lifetime, going against the never-ending nature of the task.  Indeed, the start states may never be revisited in a continuing task, so they do not deserve special treatment.  Moreover, the states experienced by a long-lived agent might not even be close to the start states, so it seems nonsensical to ask the agent to maximize short-term reward at those states.

The objective can be made more meaningful in two ways --- either by making $\mu$ representative of the states the agent will actually experience over its lifetime, or by using a longer-term future reward.  The first corresponds to setting $\mu=d_\pi$; we saw in \cref{sec:avg-reward} that $d_\pi(s)$ is proportional to how frequently the agent visits state $s$.  The second alternative corresponds to increasing the discount rate $\gamma\to1$.  We will see in \cref{sec:avg-discounted-value,sec:gamma-to-one} that both these alternatives are, in fact, equivalent to the average reward objective, and give the optimization problem
\begin{align*}
    \argmax_{\pi\in\Pi} \sum_{s\in\mathcal{S}} d_\pi(s) \, r_\pi(s)
    &\quad\equiv\quad \argmax_{\pi\in\Pi} \bar{r}(\pi), \qquad (\text{using }\eqref{eq:obj-avg-rew}) 
\end{align*}
where $\Pi$ is the set of representable policies.  This objective function captures a core tenet of reinforcement learning --- that the agent's actions should cause it to visit states where it can earn large rewards.  In continuing tasks, the agent should find policies that cause it to frequently visit these highly rewarding states.

Another alternative is to set $\mu$ to be an arbitrary weighting of states that is neither the initial state distribution nor the stationary distribution.%
\footnote{\Citet{white2017rltask} discusses weightings based on the stationary distribution but with an additional ``interest function''. For the purposes of this paper, these are generalizations of average reward.} %
If we let $\gamma\to1$, then the choice of $\mu$ is unimportant and the objective is still the average reward (see \cref{sec:gamma-to-one}). For a fixed $\gamma<1$, on the other hand, we are maximizing the short-term future reward at a set of states that must somehow be communicated to the agent (just like the reward, this is additional information that the agent would not otherwise observe).  In other words, the agent is being evaluated not on the continuing MDP, but rather on another MDP with the same state transition probabilities $P$ but a different start state distribution $\mu$.  Such an objective function could be useful in certain applications, but it represents an extension of the RL framework (similarly to off-policy objectives) and is beyond the scope of this paper.

\subsection{The average reward objective in terms of discounted values}%
\label{sec:avg-discounted-value}

Interestingly, the average reward $\bar{r}(\pi)$ has several equivalent formulations.  While we defined it as an average of the one-step reward weighted by the stationary distribution, it is equivalent to a weighted average of the discounted values:
\begin{align}
    \sum_{s\in\mathcal{S}} d_\pi(s) \, r_\pi(s)
    &\equiv (1-\gamma) \sum_{s\in\mathcal{S}} d_\pi(s) \, v_\pi^\gamma(s), &\text{for all }0\le\gamma<1. \label{eq:equivalence}
\end{align}
Intuitively, if the initial state is sampled from $d_\pi$, then all future states will have the same distribution and the expected reward on every future time step will be $\bar{r}(\pi)$.  Effectively, the stationary distribution allows any weighted average over time (in this case, the normalized value function $(1-\gamma)v_\pi^\gamma(s)$) to be replaced with a weighted average over states \citep{Singh94_POMDP_AvgRew,RLtextbook}. Indeed, the same equivalence holds for any state-independent temporal discounting scheme (not just geometric) as long the discount factors sum to a constant; the $(1-\gamma)$ factor is replaced by the reciprocal of that constant.  This equivalence does not hold for hyperbolic discounting \citep{Fedus19_HyperbolicDisc}, for example, because the sum of discount factors diverges.

Theoretically, the definition in~\eqref{eq:equivalence} implies that the policy that maximizes the average of future discounted values (weighted by the stationary distribution, but with any discount factor) also maximizes average reward.  The discount rate thus becomes a hyper-parameter of the algorithm, rather than a parameter specifying the optimization objective \citep{RLtextbook}.

\subsubsection*{Greedily maximizing discounted future value does not maximize average reward}

At this point, it would be easy to conclude that any algorithm that maximizes discounted value must also maximize average reward \emph{regardless of the discount factor}, which seems absurd on the face of it.  We should indeed be skeptical --- common algorithms like Q-learning or Sarsa estimate an action-value function $Q^\gamma(s, a)$ and behave (close to) greedily with respect to it. Such a local greedification will not in general correspond to maximizing the average discounted value over time.

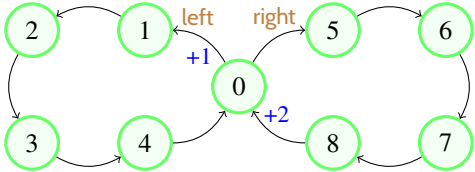
\begin{figure}[h]
    \centering
        \begin{tikzpicture}[
        roundnode/.style={circle, draw=green!60, fill=green!5, very thick, minimum size=7mm},
        ]
            \node[roundnode] at (0, 0)   (state0) {0};
            \node[roundnode] at (-1.25, 0.75)   (state1) {1};
            \node[roundnode] at (-2.75, 0.75)   (state2) {2};
            \node[roundnode] at (-2.75, -0.75)   (state3) {3};
            \node[roundnode] at (-1.25, -0.75)   (state4) {4};
            \node[roundnode] at (1.25, 0.75)   (state5) {5};
            \node[roundnode] at (2.75, 0.75)   (state6) {6};
            \node[roundnode] at (2.75, -0.75)   (state7) {7};
            \node[roundnode] at (1.25, -0.75)   (state8) {8};
        
            \draw[->] (state0) [out=120,in=0] to node[pos=0.1,left]{\small\color{blue}+1} node[pos=0.65,above]{\small\color{brown}\sffamily \ left} (state1);
            \draw[->] (state1) [out=150,in=30] to (state2);
            \draw[->] (state2) [out=240,in=120] to (state3);
            \draw[->] (state3) [out=330,in=210] to (state4);
            \draw[->] (state4) [out=0,in=240] to (state0);
            \draw[->] (state0) [out=60,in=180] to node[pos=0.5,above]{\small\color{brown}\sffamily right} (state5);
            \draw[->] (state5) [out=30,in=150] to (state6);
            \draw[->] (state6) [out=300,in=60] to (state7);
            \draw[->] (state7) [out=210,in=330] to (state8);
            \draw[->] (state8) [out=180,in=300] to node[pos=0.95,right]{\small\color{blue}+2} (state0);
        \end{tikzpicture}
        \caption{\label{fig:two-choice-MDP} The two-choice MDP\@. This is a continuing MDP with only one action in every state except state $0$. Action \textsf{left} in state $0$ gives an immediate reward of $+1$ and action \textsf{right} leads to a delayed reward of $+2$ after five time steps.}
\end{figure}

As an example, consider the two-choice MDP in \cref{fig:two-choice-MDP}. State $0$ is the only state with a choice between two actions: \textsf{left} and \textsf{right}. Choosing \textsf{left} leads to an immediate reward of $+1$, and \textsf{right} leads to a delayed reward of $+2$. For small discount factors, the immediate reward from going \textsf{left} appears much more appealing than the discounted delayed reward. When the discount factor is increased sufficiently, the \textsf{right} action becomes more appealing instead. For algorithms based on the local greedification operator, the discount factor is not just a hyper-parameter of the algorithm --- it actually changes the problem being solved.

Acting greedily with respect to a (discounted) value function is ultimately a solution method, not a problem definition. We have previously argued that the discounted objective does not define a meaningful optimization problem under function approximation. Although we cannot rule out the possibility that discounting may form part of an algorithm for control in continuing tasks,%
\footnote{For example, an algorithm might estimate $v_\pi^\gamma$ and maximize the right-hand side of the equivalence~\eqref{eq:equivalence}. Because of that equivalence, $\gamma$ would simply be a tuning parameter of the algorithm. It would have no effect on the objective being maximized, average reward. \citep[\S10.4]{RLtextbook}} %
the danger lies in naively transplanting algorithms from the episodic discounted setting to continuing tasks just because the discounted future rewards are finite in both cases. Discounting should not be the only (or even the first) algorithmic approach we consider when we set out to solve continuing RL tasks.

\subsection{Increasing $\gamma
  \texorpdfstring{\rightarrow}{ \textrightarrow} 1$ does not solve the problem}%
\label{sec:gamma-to-one}

Inspired by the previous example, one might consider using a discounted algorithm but taking the limit as $\gamma\to1$. Depending on how this limit is interpreted, however, it is either equivalent to the average reward or algorithmically impractical.

Consider an optimal policy $\argmax_\pi \lim_{\gamma\to1^{-}} (1-\gamma)v_\pi^\gamma(s_0)$. This objective does indeed avoid the issues that come with a finite (discounted) time horizon.  Furthermore, it is bounded, being the weighted average of bounded rewards.  Unsurprisingly, \citet{Bishop14_AbelCesaroLimits2014} show that this quantity is equivalent to average reward; it is independent of the start state $s_0$.

Unfortunately, that objective function contains a limit, making it hard to optimize in practice.  Instead, algorithms find optimal policies for increasing discount factors ---  they exchange the maximization with the limit --- $\lim_{\gamma\to1^{-}} \argmax_\pi (1-\gamma)v_\pi^\gamma(s_0)$ for some start state $s_0$.

In theory, the limiting policy does indeed exist and maximizes average reward; it is called the \emph{Blackwell-optimal} policy $\pi^*$.  In fact, this policy satisfies the optimality criterion~\eqref{eq:opt-policy} for every state $s$ as long as $\gamma\ge\gamma^*$, a critical discount rate that depends on the particular MDP.  Intuitively, in any given MDP, making a decision that is optimal over a sufficiently large horizon is equivalent to optimizing average reward. Crucially, this horizon depends on the MDP and how long it takes for actions to have consequences (in terms of reward), which is not known beforehand. The critical discount rate is related to the maximum mixing time of the MDP\@. \citep{Puterman94_MDPbook}

There are two major obstacles to exploiting this theory to use discounted-value algorithms as average-value algorithms for real-world RL problems.  First, most algorithms that learn discounted value functions become increasingly unstable as $\gamma$ increases to $1$.  Second, it is hard to estimate the critical discount factor, because it is intimately tied to the unknown dynamics of the environment. \Citet{Denis19_DiscountingIssues} discusses these issues more thoroughly.  Thus, to ensure we find an optimal policy we cannot settle for any fixed $\gamma<1$, forcing us into the $\gamma\approx1$ regime where our algorithms become unusable. Algorithms that optimize the average reward objective do not rely on a discount factor and automatically make decisions at the appropriate time horizon.

\subsection{Maximizing average reward is algorithmically feasible}

The average reward formulation has been long studied, and there are several dynamic programming algorithms for finding optimal average reward policies; see \citet{Puterman94_MDPbook} and \citet{Bertsekas05_DPbook}. \Citet{Schwartz93} proposed the first reinforcement learning algorithm to maximize the undiscounted average reward, variants of which were reviewed by \citet{Mahadevan96_ARsurvey}. Of particular interest is RVI~Q-learning \citep{AbounadiBB01_RVI-Q}, a simple Q-learning-like algorithm with convergence guarantees. Furthermore, policy gradient methods are especially promising since the gradient of the average reward objective has an elegant closed-form expression \citep{sutton2000policy}. \Citet{Konda99_ARPG} have proposed an actor-critic method for average reward.


\section{Conclusions}%
\label{sec:conclusions}


Discounting in continuing reinforcement learning tasks raises serious conceptual problems that become especially acute in the presence of function approximation. The key messages from this paper are as follows: 

\begin{itemize}[leftmargin=1.5em]
    \item \textbf{Discounted reinforcement learning is not an optimization problem.} In continuing tasks, the usual formulation does not correspond to the maximization of any objective function over a set of policies.
    \item \textbf{Function approximation is therefore incompatible with discounting.} Not being an optimization problem, the best representable policy is not well-defined in continuing tasks.
     \item \textbf{Average reward reinforcement learning is an optimization problem.} There is always an optimal representable policy even with function approximation.  
    \item \textbf{Greedily maximizing discounted future value does not maximize average reward.} Common algorithms like Sarsa or Q-learning do not optimize the average reward, and they find different policies depending on the discount factor.
    \item \textbf{Increasing $\gamma\to1$ does not solve the problem.}
    The discount factor must be increased beyond a critical point which is problem-specific and cannot be determined a priori. Moreover, most algorithms become increasingly unstable as $\gamma$ approaches unity.
\end{itemize}

There are many open problems in average reward RL\@. While several relatively simple convergent algorithms exist, further advances are needed in fundamental model-free and model-based learning as well as topics like multi-step learning, off-policy learning, and options. Many of the recent advances in optimization can be applied to improve algorithms for average reward.


\section*{Acknowledgements}

The authors were supported by the Natural Sciences and Engineering Research Council of Canada (NSERC), Alberta Innovates, and DeepMind. The authors also wish to thank Muhammad Zaheer, Kirby Banman, Samuel Sokota, and Martha White for valuable feedback on earlier drafts of the work.


\bibliographystyle{custom}
\providecommand{\eprint}[2][]{\href{http://arxiv.org/abs/#2}{#2}}
\def\UrlFont{\sffamily\small}
\bibliography{bibliography.bib}


\end{document}